\documentclass[sigconf]{acmart}

\usepackage{booktabs}
\usepackage{caption}
\usepackage{multirow}

\newcommand\etal{\textit{et al.}}
\newcommand\eg{\textit{e.g.}}
\newcommand\ie{\textit{i.e.}}
\newcommand\etc{\textit{etc.}}

\AtBeginDocument{%
  \providecommand\BibTeX{{%
    \normalfont B\kern-0.5em{\scshape i\kern-0.25em b}\kern-0.8em\TeX}}}
\renewcommand\footnotetextcopyrightpermission[1]{}

\acmConference[ACM MM'23]{ACM MM'23: ACM International Conference on Multimedia}{Oct.29--Nov.3, 2023}{Ottawa, Canada}

\begin{document}

\title{Physics-Based Adversarial Attack on Near-Infrared Human Detector for Nighttime Surveillance Camera Systems}


\author{Muyao Niu}
\email{muyao.niu@gmail.com}
\affiliation{%
  \institution{The University of Tokyo}
  \country{}
}

\author{Zhuoxiao Li}
\email{lizhuoxiao@g.ecc.u-tokyo.ac.jp}
\affiliation{%
  \institution{The University of Tokyo}
  \country{}
}

\author{Yifan Zhan}
\email{zhan-yifan@g.ecc.u-tokyo.ac.jp}
\affiliation{%
  \institution{The University of Tokyo}
  \country{}
}

\author{Huy H. Nguyen}
\email{nhhuy@nii.ac.jp}
\affiliation{%
  \institution{National Institute of Informatics}
  \country{}
}

\author{Isao Echizen}
\email{iechizen@nii.ac.jp}
\affiliation{%
  \institution{National Institute of Informatics}
  \country{}
}

\author{Yinqiang Zheng}
\email{yqzheng@ai.u-tokyo.ac.jp}
\affiliation{%
  \institution{The University of Tokyo}
  \country{}
}

\begin{abstract}

  Many surveillance cameras switch between daytime and nighttime modes based on illuminance levels. During the day, the camera records ordinary RGB images through an enabled IR-cut filter. At night, the filter is disabled to capture near-infrared (NIR) light emitted from NIR LEDs typically mounted around the lens. While RGB-based AI algorithm vulnerabilities have been widely reported, the vulnerabilities of NIR-based AI have rarely been investigated. In this paper, we identify fundamental vulnerabilities in NIR-based image understanding caused by color and texture loss due to the intrinsic characteristics of clothes' reflectance and cameras' spectral sensitivity in the NIR range. We further show that the nearly co-located configuration of illuminants and cameras in existing surveillance systems facilitates concealing and fully passive attacks in the physical world. Specifically, we demonstrate how retro-reflective and insulation plastic tapes can manipulate the intensity distribution of NIR images. We showcase an attack on the YOLO-based human detector using binary patterns designed in the digital space (via black-box query and searching) and then physically realized using tapes pasted onto clothes. Our attack highlights significant reliability concerns for nighttime surveillance systems, which are intended to enhance security. Codes Available: \url{https://github.com/MyNiuuu/AdvNIR}.

\end{abstract}





\maketitle


\section{Introduction}

Surveillance camera systems are widely deployed for security purposes, capturing ordinary RGB images during the daytime while filtering out the infrared component with an IR-cut filter. However, in extremely low-light conditions such as midnight, clear visible images are difficult to obtain due to limited camera sensitivity. To address this, NIR LEDs are often utilized for auxiliary illumination.
The central wavelength of these NIR LEDs is typically around 850nm, falling within the spectral response range of silicon-based image sensors. This NIR light is invisible to human eyes and helps to reduce light pollution while concealing surveillance coverage.

With the rapid development of deep learning-based visual AI, surveillance cameras are becoming more intelligent, with advanced functionalities of human detection~\cite{redmon2016you,carion2020end}, face recognition~\cite{wen2016discriminative,schroff2015facenet}, low-light enhancement~\cite{niu2023visibility, niu2023nir}, \etc. The vulnerability of RGB-based vision has also been extensively studied, and countermeasures to protect them from attack have been proposed~\cite{hu2022adversarial,duan2021learning,huuu2022adversarial,suryanto2022dta,wang2022fca,tan2021legitimate}. 

However, the potential vulnerabilities of NIR-based vision have rarely been explored. Although there have been attempts to use electronic devices worn on the face that emits near-infrared lights to hide human faces from cameras~\cite{yamada2013privacy,yamada2012use}, the potential vulnerabilities of NIR-based human body detectors have never been reported in existing studies. Given its wide application scenarios in security fields (\eg, surveillance cameras), these vulnerabilities may leave a window for attackers to exploit, causing great security issues. 


In this paper, we bring this critical concern to the community and warn of the potential risks of using AI algorithms in the NIR domain~\cite{redmon2016you,wu2017rgb,zhang2022fmcnet} for existing surveillance systems. By considering the imaging principle of NIR nighttime surveillance camera systems and the installation method of auxiliary lighting, several inherent vulnerabilities make NIR-based AI more brittle and less robust compared to RGB-based AI.

\begin{figure}[t]
\centering
\includegraphics[width=\linewidth]{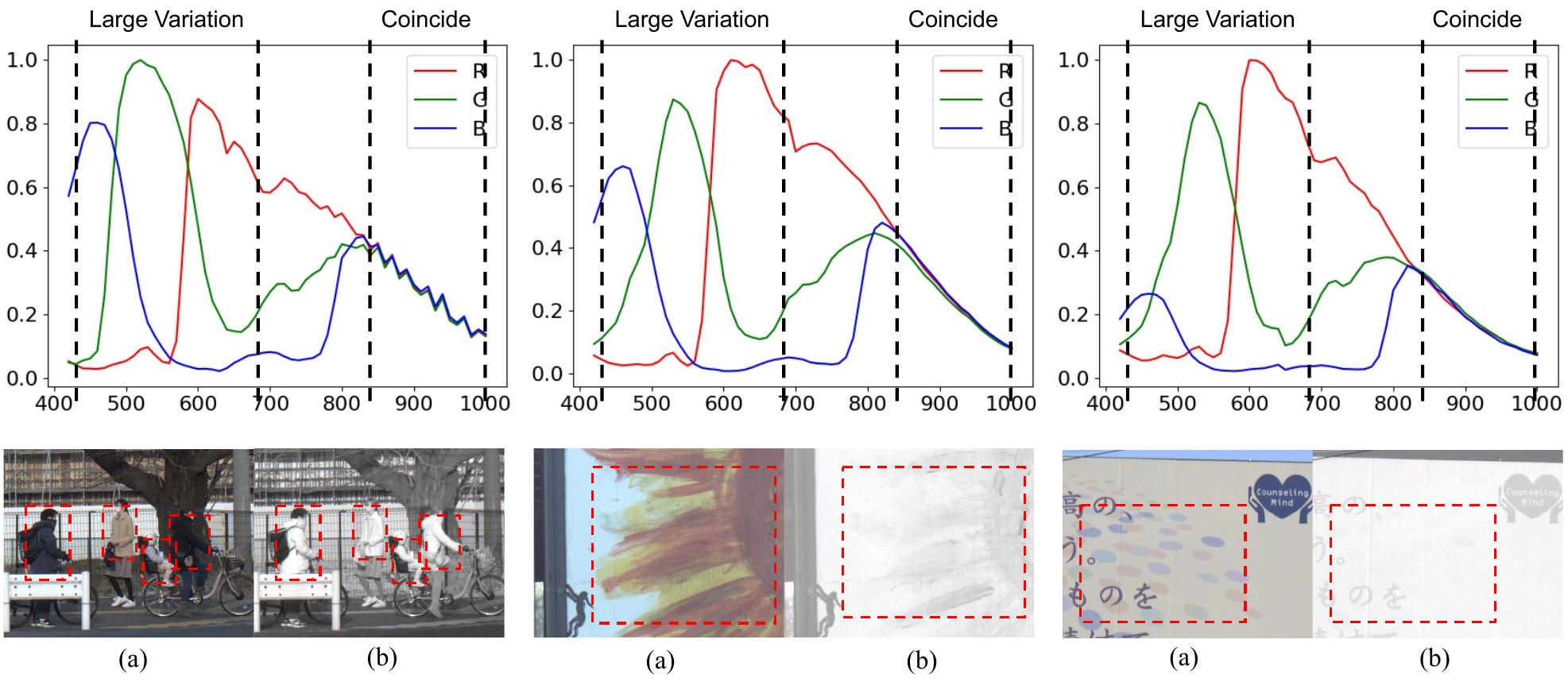}
\setlength{\abovecaptionskip}{-4mm}
\caption{Illustration for color loss in NIR. Up: Camera Spectral Sensitivities of three different color cameras. Down: RGB images (a) and NIR images (b) for the same scene.}
\label{fig:colorloss}
\vspace{-4mm}
\end{figure}

Specifically, compared to visible illuminations, the NIR LED illumination around 850 nm has the disadvantage that the color and texture of scene objects captured by the camera tend to disappear (Fig.~\ref{fig:colorloss}),
and this phenomenon particularly happens on textures of dyed fabrics such as human clothing~\cite{wu2020learn,liu2022optimal}. The reason lies in the fundamental imaging principle in the NIR range, which can be divided into two aspects. First, the spectral sensitivities of the three color filters of the silicon camera tend to overlap with each other in the NIR region, so monochromatic images can be acquired (Fig.~\ref{fig:colorloss}),
though the sensor itself is trichromatic and the scene has brilliant colors. Second, the reflectance spectra of different materials also tend to coincide, resulting in severe texture loss. As a result, the intensity variation of NIR images is very limited, and the robustness of AI models trained on such data tends to be unstable. This implies the immediate vulnerability of AI models when the intensity distribution of the input image is somehow altered.

\begin{figure}[t]
\centering
\includegraphics[width=\linewidth]{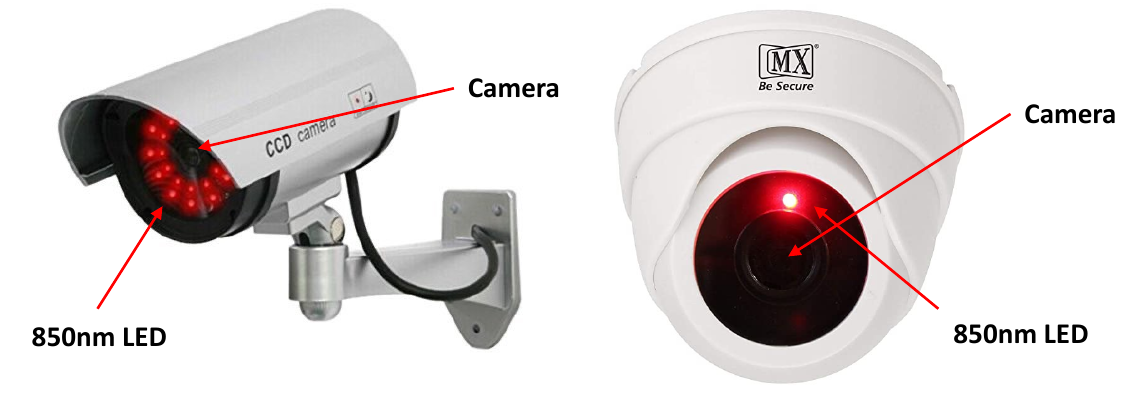}
\setlength{\abovecaptionskip}{-2mm}
\caption{Co-located setup of surveillance camera systems.}
\label{fig:colocated}
\vspace{-4mm}
\end{figure}

Regarding how to easily manipulate the intensity distribution in the physical world, we further recognize an intrinsic limitation of the geometric location of the auxiliary illuminant and the camera. To save installation space and reduce occlusion, LEDs are usually mounted around the lens (Fig.~\ref{fig:colocated}), which leads to a nearly co-located geometry setup. Given such a setting, the scene surface perpendicular to the viewing/lighting direction will lead to saturation when encountering certain materials. 


In reality, we propose to use retro-reflective tape, which by design has a micro surface perpendicular to the viewing/lighting direction, irrespective of the geometric shape of the tape. This can be used to brighten an image area when needed. On the other hand, black plastic tapes (e.g., electrical insulating tape) have low NIR reflectance, allowing to reduce the intensity of an image area. Since retro-reflective tape's visible color (to human eyes) is independent of its reflectivity in the NIR range, we can use different color combinations to make the physical attacker appear natural to human eyes, further improving its stealthiness.

Based on the aforementioned vulnerabilities of NIR and the intensity manipulation via tapes with different materials, we showcase an attack instance on the YOLO-based human detector by first designing binary patterns in the digital space via black-box query and searching, neither requiring knowing the model parameters nor using any gradient-based optimization techniques. We leverage the 3D human models and render-er during the design process to consider real-world physical attributes. As a result, the designed pattern can be physically realized with appropriate tapes pasted onto clothes whose attack ability has been verified. Since retro-reflective tapes and insulation tapes are easily available at very low prices and the cost to make a physical attacker is extremely low, the proposed method will pose tremendous reliability concerns on NIR-based AI, widely introduced in nighttime surveillance systems for security and convenience. 

The contribution of this work can be summarized as follows:
\begin{itemize}
    \item We disclose the shortage of intensity variation in the NIR domain and analyze its physical mechanism from the perspective of camera spectral response and reflectance of dyed cloth, which finally leads to fundamental vulnerabilities of AI models on such images.
    \item We propose a fully passive yet effective method to manipulate the intensity distribution of NIR images captured by a co-located camera/illuminant system, mainly grounded on the well-known principle of retro-reflective materials. 
    \item We show an instance of how to attack the deep-learning based human detector in the blackbox style. Based on the principle of retro-reflective materials, the attacker is designed in a 3D-aware manner by introducing the 3D human models and rendering techniques in the digital space and can be easily implemented in the physical space with superior generalization abilities.
\end{itemize}

\section{Related Works}

\noindent \textbf{Physical Adversarial Attacks.} The RGB-based vision models grounded on deep learning are shown to be vulnerable to adversarial attacks, and various attacking methods
have been extensively studied, which further inspire a variety of measures to protect RGB-based vision from being attacked. We refer readers to those excellent review papers for more information~\cite{moosavi2017universal,dong2018boosting,goodfellow2014explaining,madry2017towards,athalye2018synthesizing,carlini2017towards}. Instead of attacking an AI model in the digital space, more recent researches have tried to find effective attackers that can be physically realized, either by using contact-free illumination or physical attackers that should be pasted in the scene or on the target object~\cite{zhong2022shadows,zolfi2021translucent,jan2019connecting,eykholt2018robust,duan2020adversarial,duan2021adversarial,liu2019perceptual,liu2020bias,Zolfi_2021_CVPR}. 
Among these methods, Whitebox attacks~\cite{wu2020making,xu2020adversarial,tan2021legitimate,hu2021naturalistic,xingxing2023physically} utilize the training data and possess knowledge of the model, including its structure, parameters, weights, \etc. Consequently, these attacks are relatively straightforward to implement. 
Wu~\etal~\cite{wu2020making} and Xu~\etal~\cite{xu2020adversarial} used physically printed posters and wearable clothes to realize patch-based attacks, \eg, adversarial T-shirt with designed attack patterns. However, these designed patterns are usually not stealthy enough due to its unnatural patterns. To address this problem, several works are proposed to generate cartoon-like patches that look more natural~\cite{tan2021legitimate,hu2021naturalistic}. On the other hand, Blackbox attacks~\cite{zhong2022shadows,wei2022hotcold,diao2021basar,li2021adversarial,wei2020heuristic} involve situations where the attacker has no access to the structure and parameters of the target model. 
Zhong \etal~\cite{zhong2022shadows} proposed a light-based blackbox attack via the very common natural phenomenon—shadow. In this paper, we exploit the Blackbox setting which have wider application scenarios since it does not need to know the structure and parameters of the target model.


\noindent \textbf{Adversarial Attacks beyond RGB.} 
The majority of researches of adversarial attacks focus on RGB images, and the vulnerabilities of visual AI based on other spectral modalities are less explored. 
Zhu~\etal~\cite{zhu2021fooling} proposed an adversarial attack method that creates thermal infrared patterns using glowing light bulbs. Later, they further designed an infrared invisible clothing using aerogel material to fool state-of-the-art human detectors~\cite{Zhu_2022_CVPR}. 
For blackbox adversarial attack, HotCold Block~\cite{wei2022hotcold} was recently proposed to perform thermal infrared attack by using warming paste and cooling paste. 

To attack NIR face detection, Yamada~\etal~\cite{yamada2013privacy,yamada2012use} use electronic devices worn on the face that emits near-infrared lights to hide human faces from being identified. However, it is not wise to extend such an active attack method to hide full body from cameras. Our method is fully passive grounded on the physical characteristics of modern surveillance camera systems, and it attacks advanced human detectors. 
To our knowledge, no research has been conducted to disclose the vulnerability of NIR-based visual AI for human body detection, in spite of the fact that this is directly relevant to nighttime surveillance systems widely installed for daily usage. 

\section{Vulnerabilities of NIR Photography}
\label{sec:vul}

\begin{figure*}[t]
\centering
\includegraphics[width=\textwidth]{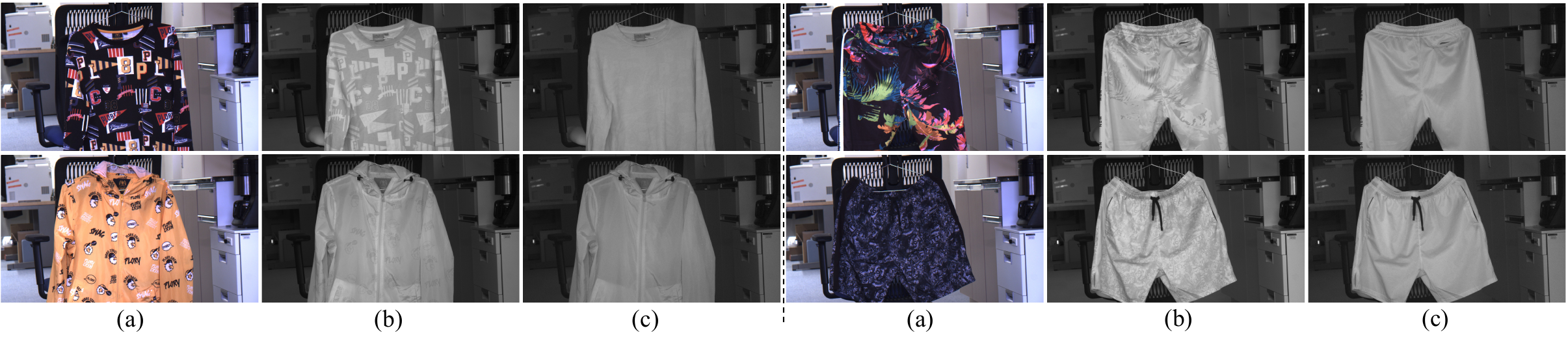}
\setlength{\abovecaptionskip}{-2mm}
\caption{Illustration for texture loss in NIR. (a) RGB images with IR-cut filter, (b) NIR images through $>$ 700nm long-pass filter, (c) NIR images through 850nm band-pass filter with a full width at half magnitude (FWHM) of 40nm, whose spectral response is very close to the spectral distribution of 850nm LEDs. All images are captured with the same trichromatic camera, with its IR-cut filter removed when needed (b,c). We can see that, although the clothes have rich color and texture in the visible range, they tend to disappear, especially in the narrow spectral range covered by 850nm LEDs.}
\label{fig:textureloss}
\vspace{-2mm}
\end{figure*}

\begin{figure}[t]
\centering
\includegraphics[width=\linewidth]{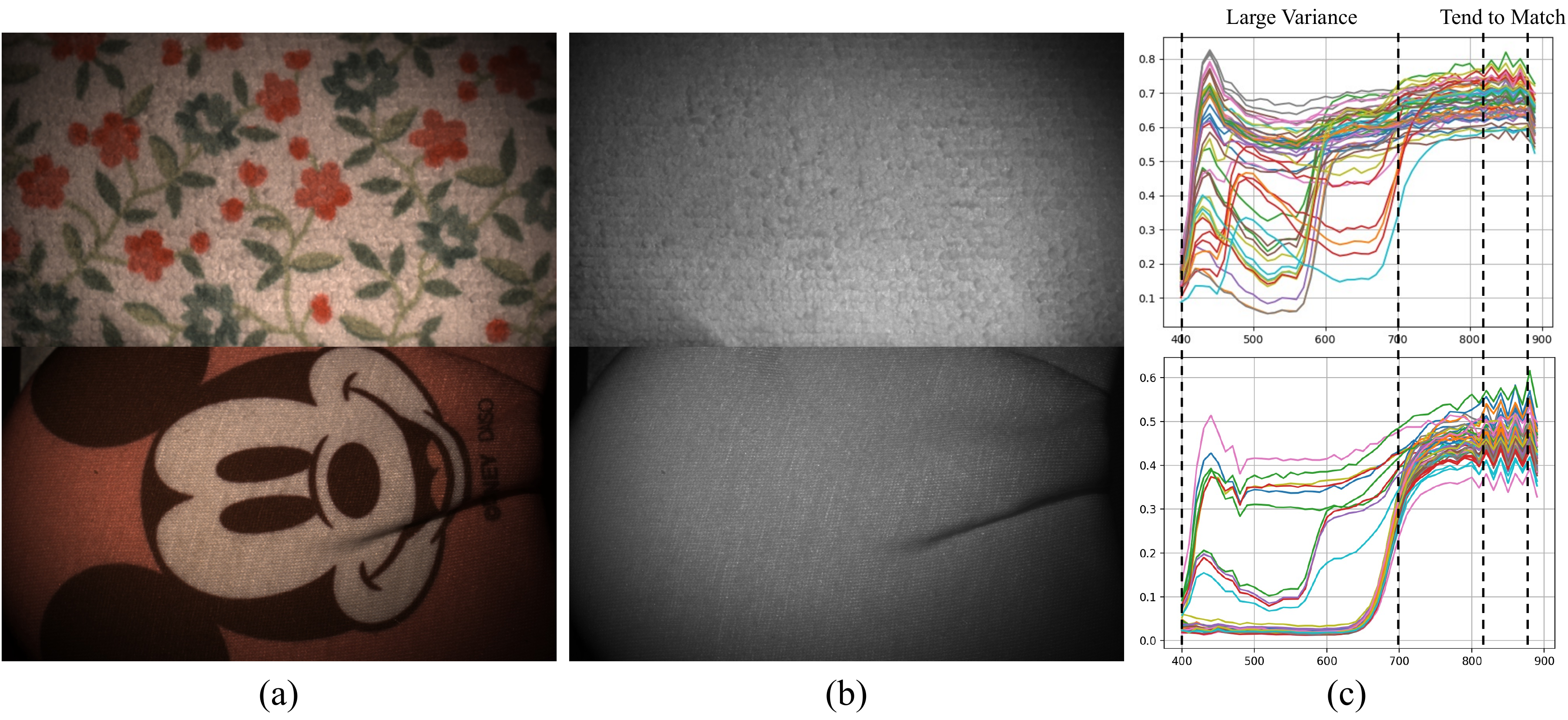}
\setlength{\abovecaptionskip}{-2mm}
\caption{Analysis on texture loss in NIR. (a) RGB images, (b) 850nm NIR images. (c) 50 Spectral reflectance sampled from each scene. We can see that the spectral reflectances of different scene points tend to coincide around 850nm. }
\label{fig:textureloss2}
\vspace{-2mm}
\end{figure}

\begin{figure}[t]
\centering
\includegraphics[width=\linewidth]{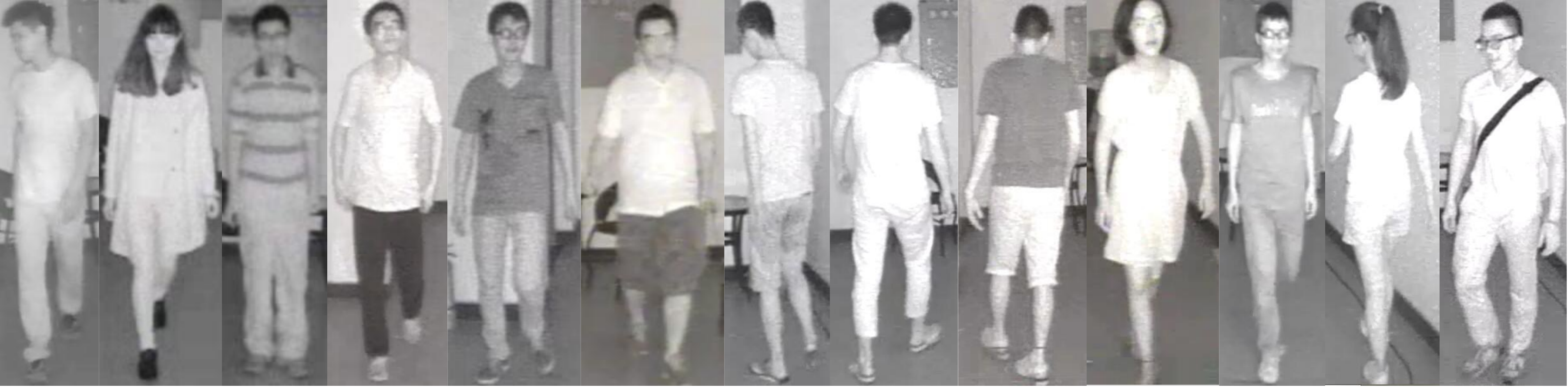}
\caption{Examples of color and texture loss in human clothes from public datasets~\cite{wu2017rgb}.}
\label{fig:textureloss_SYS}
\vspace{-4mm}
\end{figure}

In this section, we analyze the fundamental vulnerabilities of NIR surveillance from three aspects: color loss, texture loss, and co-located settings. These three vulnerabilities allow our fully passive yet effective attack in both the digital world and the physical world, which will be further described in Sec.~\ref{sec:method}.

\noindent \textbf{Color loss.} For dual-mode surveillance cameras, NIR images are captured by trichromatic sensors since they also have spectral sensitivities in the NIR range. However, different from RGB images, NIR images appear to be monochromatic. The fundamental reason is that the Camera Spectral Sensitivity (CSS) in the NIR wavelengths is almost the same for Red, Green, and Blue channels. As shown in Fig.~\ref{fig:colorloss}, the intensity of three channels coincide in the NIR range, leading to gray-scale output. As a result, the distribution of NIR images is simpler, with lower color variations compared to RGB images due to color loss. 

\noindent \textbf{Texture loss.} In addition to color loss, the texture loss further makes NIR data distribution more monotonic. Fig.~\ref{fig:textureloss} shows three types of images taken by the same color camera through different filters. We can see that the texture of cloth, dye, and fabrics gradually disappears in the NIR ranges ($>$ 700nm), and the texture loss becomes even worse when only using light around 850nm. The fundamental reason lies in the spectral reflectance of these materials. Fig.~\ref{fig:textureloss2} shows two typical examples. We can see that the spectral reflectance intensity of different scene points varies a lot in the RGB ranges, but tends to be overlapped in the NIR range, especially around 850nm. Thus, these materials can produce different RGB observations in the visual range but have almost the same appearance in an 850nm NIR image. 

Given the severe color and texture loss in 850nm NIR images, it is observed that clothes worn by people have rather simple distributions than RGB images -- uniform white with no texture. Fig.~\ref{fig:textureloss_SYS} shows some examples from the dataset collected by~\cite{wu2017rgb}. We argue that this makes an AI model trained on these data less robust and can easily fail when encountering more complex or different data distributions.

\begin{figure*}[t]
\centering
\setlength{\abovecaptionskip}{0mm}
\includegraphics[width=\linewidth]{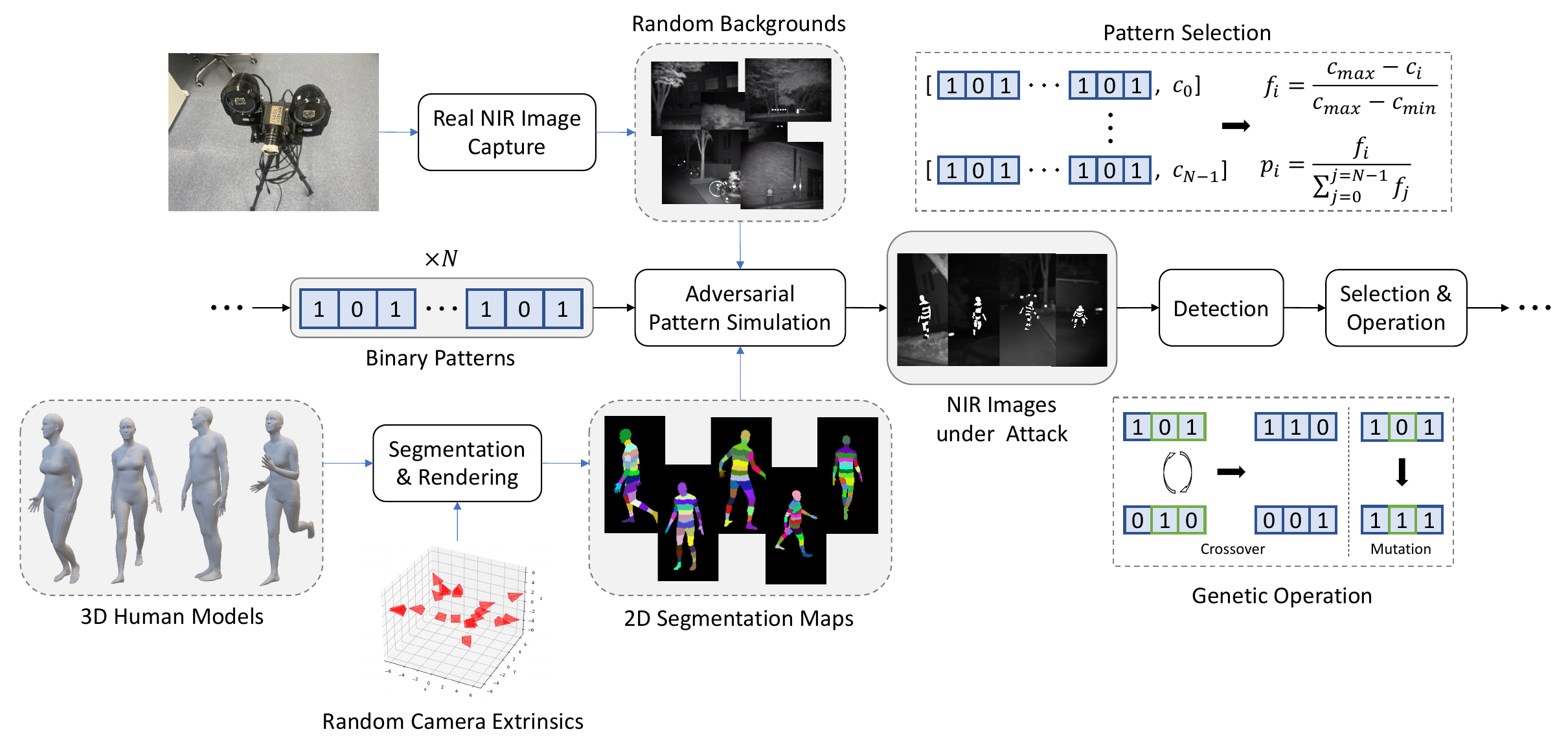}
\caption{Demonstration of one iteration for our pipeline. During each iteration, we update $N$ groups of binary patterns based on the detection results of the corresponding attacked images. The process involves sampling a batch of 3D human models and segmenting each of them into $K$ parts. We then generate 2D segmentation maps by rendering the models with arbitrary camera positions. Each part of the segment map is assigned a value of either $0$ or $255$ based on the corresponding binary pattern. We combine these maps with NIR backgrounds captured by a real camera system to generate the attacked NIR images. The detector then uses these images as input to produce the detection results. Finally, we update each binary pattern based on the detection results using selection and genetic operations.}
\label{fig:pipeline}
\vspace{-4mm}
\end{figure*}

\noindent \textbf{Co-located settings.} Given the simple and vulnerable distributions of 850nm NIR images, effective attacks can be achieved by manipulating the light distribution of NIR images. We notice that surveillance camera systems always have NIR LEDs mounted around the lens (Fig.~\ref{fig:colocated}), for the sake of saving installation space and avoiding occlusion. However, this co-located setting may easily lead to saturation and over-exposure when the direction of the light beam emitted from the LED matches the direction of the light beam entering the lens. Furthermore, if the camera’s auto-exposure function is activated to reduce the saturation effect, other image areas are likely to be darkened. 
Given these characteristics of NIR-based surveillance, we believe that it is possible to easily manipulate the local luminance distribution of images for an effective attack.

\section{Method}
\label{sec:method}
We first introduce how we control local light intensity for an effective attack using two types of tapes. We then introduce the composition of the designed adversarial attacker in the digital space and finally describe our proposed adversarial searching algorithm on how to find an effective adversarial patterns.

\subsection{Problem Definition}

Given an NIR image $I \in \mathbb{R}^{3 \times H \times W}$ where $H$ an $W$ are respectively the height and width of the input image, the pre-trained human detector $F$ will return an output label $\mathcal{Y}$ that includes the position of the bounding boxes $Y_{pos}$, and the corresponding confidence $Y_{conf}$:
\begin{align}
    \mathcal{Y} = F(I) = [Y_{pos}, Y_{conf}].
\end{align}
Our goal is to fool the human detector and make it fail to detect humans in the images, \ie, minimizing the $Y_{conf}$:
\begin{align}
    \arg \min Y_{conf}.
\end{align}
To achieve this, we design an attack algorithm $f_{atk}$ that takes an NIR image $I$ as input and outputs the attacked image $\hat{I}$ for which the human detector has the lowest confidence:
\begin{align}
    \arg \min F(f_{atk}(I)).
\end{align}

\subsection{Manipulating Light Intensity with Tapes}

\begin{figure}[t]
\centering
\includegraphics[width=0.8\linewidth]{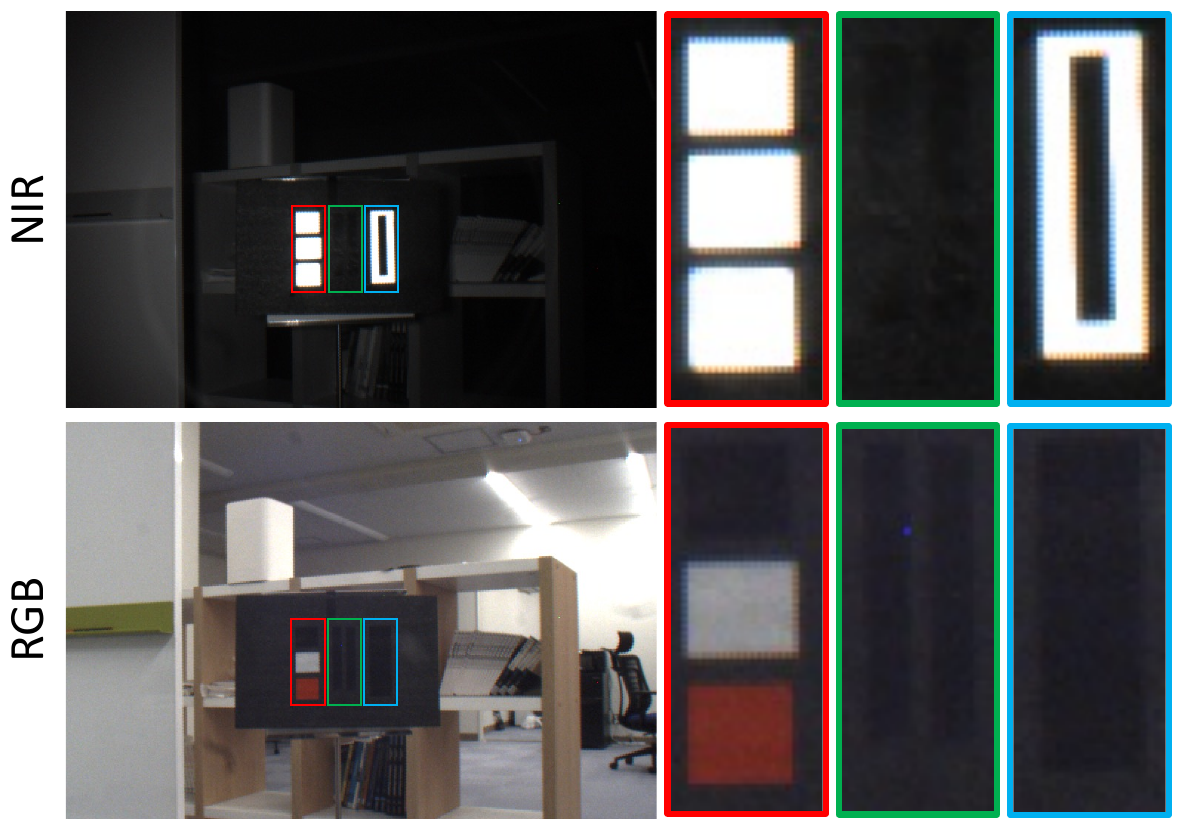}
\setlength{\abovecaptionskip}{2mm}
\caption{Performance of retro-reflective tape and insulating tape, when captured by a camera with co-located 850nm LED. Red: retro-reflective tape. Green: insulating tape. Blue: insulating tape pasted to a bigger retro-reflective tape. We can see that the retro-reflective tapes are always bright, irrespective of their visible colors.}
\label{fig:tapes}
\vspace{-4mm}
\end{figure}

Using the feature of surveillance camera systems mentioned in Sec.~\ref{sec:vul}, we consider two types of tapes that can easily manipulate the local intensity of NIR images: Black plastic tapes (\eg, electrical insulating tape) have low NIR reflectance, enabling them to reduce the luminance value of an NIR image. Retro-reflective tapes, on the other hand, are made based on the retroreflective principle which can increase the luminance value because the direction of the light beam emitted from the LED and the direction of the light beam entering the lens are always the same, regardless of the orientation of the tape. By using these two types of tapes, we can easily control the local intensity of an NIR image to either black or white. Fig.~\ref{fig:tapes} show the performance of these tapes. We can see that the tapes can easily achieve light intensity manipulation due to their optical properties as long as the light and the camera are in co-located settings. Another observation is that the performance of retro-reflective tape is independent of its visible color, and even black retro-reflective tape can equally brighten NIR images, a key advantage that allows implementing physical attackers with natural and coherent appearances to human eyes.

\subsection{Adversarial Medium Design in the Digital Space}

\label{sec:digital}

In this section, we provide a detailed explanation of our adversarial medium design pipeline, which is illustrated in Fig.~\ref{fig:pipeline}. Our approach involves updating $N$ groups of binary patterns in each iteration, where the length of each pattern is $K$. To achieve this, we begin by obtaining 3D human models and partitioning each of them into $K$ parts. We then generate 2D segmentation maps by rendering these models from various camera positions. For each part of the segmentation map, we assign a value of either $0$ or $255$ based on the corresponding binary pattern. These maps are combined with actual NIR backgrounds captured by our camera system to create the attacked NIR images. The detector uses these images as input and produces detection results. We then utilize selection and genetic operations to update the binary patterns based on the detection results obtained for each pattern.

In the following paragraphs, we will introduce each part respectively, including 3D model segmentation and rendering, in-the-wild NIR image capture, adversarial pattern simulation, detection, binary pattern selection, and genetic operation.

\subsubsection{3D model segmentation and rendering.} To perform the 3D physics-based adversarial attack, we need to mimic the real-world scenario in the digital space to assure that our result, searched in the digital space, can also generalize when adapting to the physical space. To achieve this, we adopt 3D models and rendering techniques to our pipeline. For 3D models, we use the most widely used SMPL-based~\cite{SMPL_2015,SMPL_X_2019,MANO_SIGGRAPHASIA_2017} human models to construct the 3D meshes, and randomly sample motions from the CMU Graphics Lab Motion Capture Database~\footnote{http://mocap.cs.cmu.edu/}.

\begin{figure}[t]
\centering
\setlength{\abovecaptionskip}{-1mm}
\includegraphics[width=\linewidth]{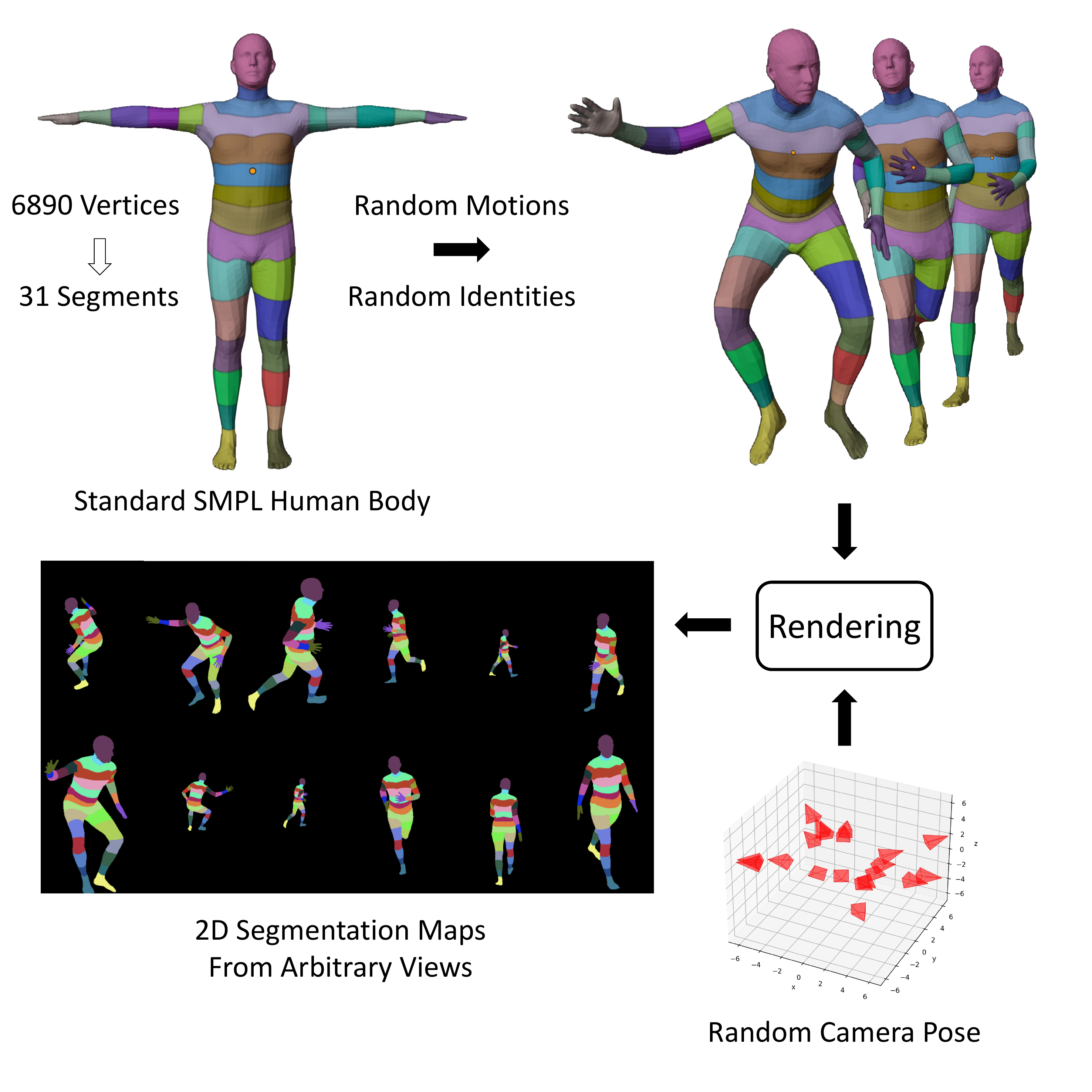}
\caption{Our segmentation paradigm on 3D human models to obtain 2D segmentation maps. Given the standard SMPL~\cite{SMPL_2015,SMPL_X_2019,MANO_SIGGRAPHASIA_2017} Human Body Model which has 6890 vertices, we segment all the vertices into 31 groups according to the semantic meanings of each part. Thanks to the homeomorphism characteristics of all meshes generatd by the SMPL-based model, the segmentation results remain consistent on different 3D models with different identities and motions. We then render these models with various camera extrinsics and obtain the 2D segmentation maps.}
\label{fig:segmentation}
\vspace{-4mm}
\end{figure}

To generate 2D segmentation maps, we need to segment each of the 3D human models into $K$ pieces. The segmentation process is shown in Fig.~\ref{fig:segmentation}. We start from the standard SMPL~\cite{SMPL_2015,SMPL_X_2019,MANO_SIGGRAPHASIA_2017} human model which has 6890 vertices and split it into $K=31$ pieces according to the semantic meanings of each part. It is worth noting that meshes generated by the SMPL model have the same number of vertices, order, and connectivities, regardless of the 3D mesh's identity or motion. This consistency ensures that our 3D segmentation results remain uniform across various 3D models. Given that attaching tapes to the head, hands, and feet of a human body can be difficult, it is beneficial to attach NIR texture instead of either black or white to the corresponding parts (a total of 5 pieces) of the 3D model. However, there is currently no research or datasets available on 3D texture modeling for these parts in the NIR domain. Considering the auto-exposure function (disussed in Sec.~\ref{sec:vul}) that has been widely adapted to the nighttime surveillance systems, we force these 5 parts to black during the design process.

Finally, we randomly sample camera poses and render the 2D segmentation maps.

\begin{figure*}[t]
\centering
\setlength{\abovecaptionskip}{0mm}
\includegraphics[width=\linewidth]{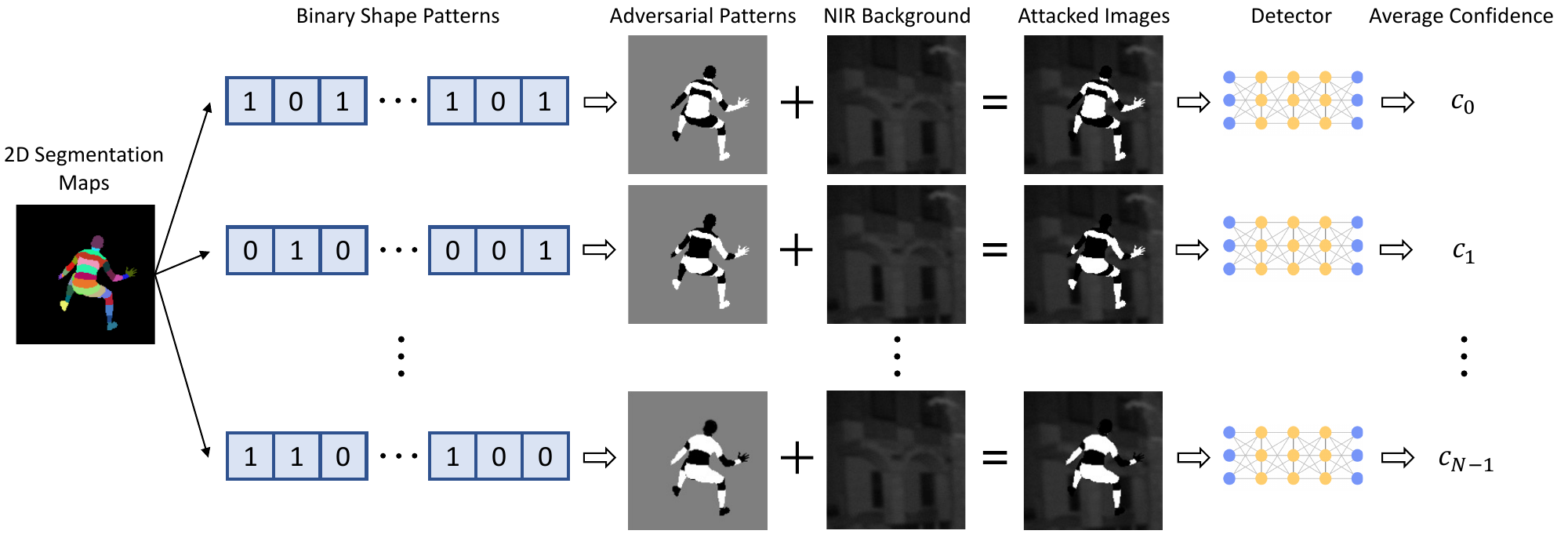}
\caption{Demonstration for the adversarial pattern synthesis. Given a batch of 2D segmentation maps rendered from various views, we assign values of $0$ or $255$ to each segment of the maps based on different binary patterns (assign $255$ when the binary pattern equals to $1$) to create the adversarial patterns. We then combine these adversarial patterns with real NIR backgrounds captured by our system to create the final attacked images. We feed these images to the detector and compute the average confidence score for each binary pattern, which we use to evaluate their rank in the subsequent selection process.}
\label{fig:adversarial}
\vspace{-2mm}
\end{figure*}

\subsubsection{Adversarial pattern simulation.}

As discussed in Sec.~\ref{sec:vul}, illumination intensities in the NIR images can be easily controlled by electrical insulating tape (low intensity, \ie, black) and retro-reflective tapes (high intensity, \ie, white). To create our adversarial pattern, we define the value of each pixel as a binary variable $x \in \{0,255\}$, with $0$ representing black and $255$ representing white. Based on the synthesized 2D segmentation maps, we assign each pixel in a given region of the map to either $0$ or $255$ based on $K$-length binary patterns that are updated during the searching process. We maintain $N$ binary patterns during the search, each of which produces a different adversarial pattern for attack, as illustrated in Fig.~\ref{fig:adversarial}.

\noindent \textbf{In-the-wild NIR background image.} Since the adversarial patterns can not simulate background contents in NIR images, we capture in-the-wild NIR images using the GS3-U3-15S5C camera with the 850nm band-pass filter to provide real NIR backgrounds for simulated attack images. For each group of adversarial patterns, we randomly sample the NIR background image during the search process.

\subsubsection{Detection.}

After obtained the simulated attacked NIR images $\hat{I} \in \mathbb{R}^{3 \times H \times W}$ where $H$ an $W$ are respectively the height and width of the image, we feed them into the pre-trained human detector $F$ and obtain the position of the bounding boxes $Y_{pos}$, and the corresponding confidence $Y_{conf}$:
\begin{align}
    F(\hat{I}) = [Y_{pos}, Y_{conf}].
\end{align}

\subsubsection{Binary pattern selection.}

Retrieving the optimal pattern without gradient from the target model is not an easy task. Here we search for the optimal adversarial pattern based on the Genetic Algorithm since our adversarial patterns are controlled by binary values. 

During each iteration, we select a portion of the existing binary patterns to reproduce for the next iteration. Each binary pattern is selected through a ranking-based process, where individuals with higher rankings (as measured by a ranking function) are more likely to be selected.

Since our goal is to fool the human detector and makes it fail to detect humans in the images, \ie, minimizing the $Y_{prob}$, we use the average confidence $c_i$ on the image batch to evaluate the fitness of each binary pattern. For each binary pattern, we calculate a ranking score $f_i$ through the ranking function:
\begin{align}
    f_i = \frac{c_{max}-c_i}{c_{max}-c_{min}},
\end{align}
where $c_{max}$ is the maximum value among all $c_i$, and $c_{min}$ is the minimum value.

These ranking scores are used to select better patterns to the next iteration. Specifically, each binary pattern has a chance of $p_i$ to be chosen for the next iteration, where $p_i$ is defined as:
\begin{align}
    p_{n}=\frac{f_n}{\sum_{j=0}^{N-1} f_j}.
\end{align}
Note that the selection process will be performed $N$ times to generate $N$ individuals for the next iteration.

\subsubsection{Genetic operation.}

Once we have selected a portion of the binary patterns through the ranking-based process, we apply traditional Genetic Algorithm operations including Crossover and Mutation. The probability of Crossover and Mutation is determined by $P_{cross}$ and $P_{mut}$ respectively. In Crossover, we randomly select two binary patterns and determine a crossover index. The parts of these two binary patterns are then exchanged from the crossover index. For Mutation, we randomly choose a mutation index and change its value (either $0 \rightarrow 1$ or $1 \rightarrow 0$).

\section{Experiments}


\subsection{Settings}


\noindent \textbf{Metrics.} We adopt two metrics to evaluate the effectiveness of our model: (1) Average Confidence (AC) for person class, which is defined as the average confidence the detector outputted for the person class. (2) Attack Success Rate (ASR) ($\%$), which is defined as follows:
\begin{align}
\operatorname{ASR} = 1 - \frac{N_a}{N_0}
\end{align}
where $N_0$ is the number of all true positive labels detected in our dataset when there is no attack, and $N_a$ is the number of all person labels detected in our dataset after attack. The higher the ASR, the more effective the adversarial attack method is.

\noindent \textbf{Implementation details.} 
We use YOLOv5~\footnote{https://github.com/ultralytics/yolov5} as our target model since it is fast, effective, and widely used. We first use YOLO5S with the official checkpoint as our attack target. We then finetune an NIR version of YOLO5S on our collected NIR dataset and use it as our attack target. Our collected dataset to finetune YOLO5S contains 13000 video frames of 13 identities with varying and cluttered backgrounds, and each identity has 500 frames of frontside and 500 frames of backside. During the searching process, the number of binary patterns is set to $N=1000$, and the image batch size is set to $B=300$. $P_{cross}$ and $P_{mut}$ are set to 0.5 and 0.01, respectively. We conduct all experiments on an NVIDIA GeForce RTX 3090 GPU, and all our codes are implemented in PyTorch~\cite{paszke2019pytorch}.

\noindent \textbf{Details of Digital Data Synthesis.} As has been mentioned in Sec.~\ref{sec:digital}, to perform pattern search and attack in digital space, we incorporate SMPL and CMU Graphics Lab Motion Capture Database to synthesize different 3D human meshes and introduce different motions to generate our human model, then segment and render these 3D meshes with different camera extrinsics. We then add random backgrounds from our taken real NIR images. To achieve this, we randomly generate 428 human meshes with different identities and motions, then segment and render each human model with 25 different camera extrinsics. These lead to 10700 different 2D segment maps. We use 8560 of them to search the adversarial pattern via our proposed algorithm and use the rest 2140 to test the effectiveness of our designed adversarial pattern. The random NIR backgrounds come from 909 NIR images taken at nighttime with camera GS3-U3-15S5C and two co-located 850nm NIR LEDs.

\begin{table}[!t]
\centering
\caption{Average Confidence (AC) and Attack Success Rate (ASR) ($\%$) results of different adversarial patterns on digital space. Best results are in \textbf{Bold}.}
\resizebox{\linewidth}{!}{
\begin{tabular}{lcccc}
\toprule
\multirow{2}{*}{Patterns} & \multicolumn{2}{c}{Finetuned YOLO} & \multicolumn{2}{c}{Official YOLO}                           \\
\cmidrule(r){2-3} \cmidrule(r){4-5}
      & AC $\downarrow$ & ASR $\uparrow$ & AC $\downarrow$ & ASR $\uparrow$ \\ 
\midrule
All black     & 0.182 $\pm$ 0.002 & 75.94 $\pm$ 0.68 & 0.853 $\pm$ 0.001 & 0.72 $\pm$ 0.18 \\
All white     & 0.262 $\pm$ 0.003 & 58.41 $\pm$ 0.57 & 0.567 $\pm$ 0.004 & 23.23 $\pm$ 0.58 \\
Random        & 0.160 $\pm$ 0.030 & 79.80 $\pm$ 6.88 & 0.655 $\pm$ 0.083 & 12.88 $\pm$ 7.99 \\
Ours (5black)          & 0.098 $\pm$ 0.002 & 93.44 $\pm$ 0.49 & 0.370 $\pm$ 0.004 & 46.64 $\pm$ 0.84 \\
Ours (1black)          & \textbf{0.097 $\pm$ 0.001} & \textbf{94.14 $\pm$ 0.45} & \textbf{0.301 $\pm$ 0.004} & \textbf{56.23 $\pm$ 1.11} \\

\bottomrule
\end{tabular}
}
\label{tab:digital}
\end{table}

\begin{figure}[t]
\centering
\includegraphics[width=\linewidth]{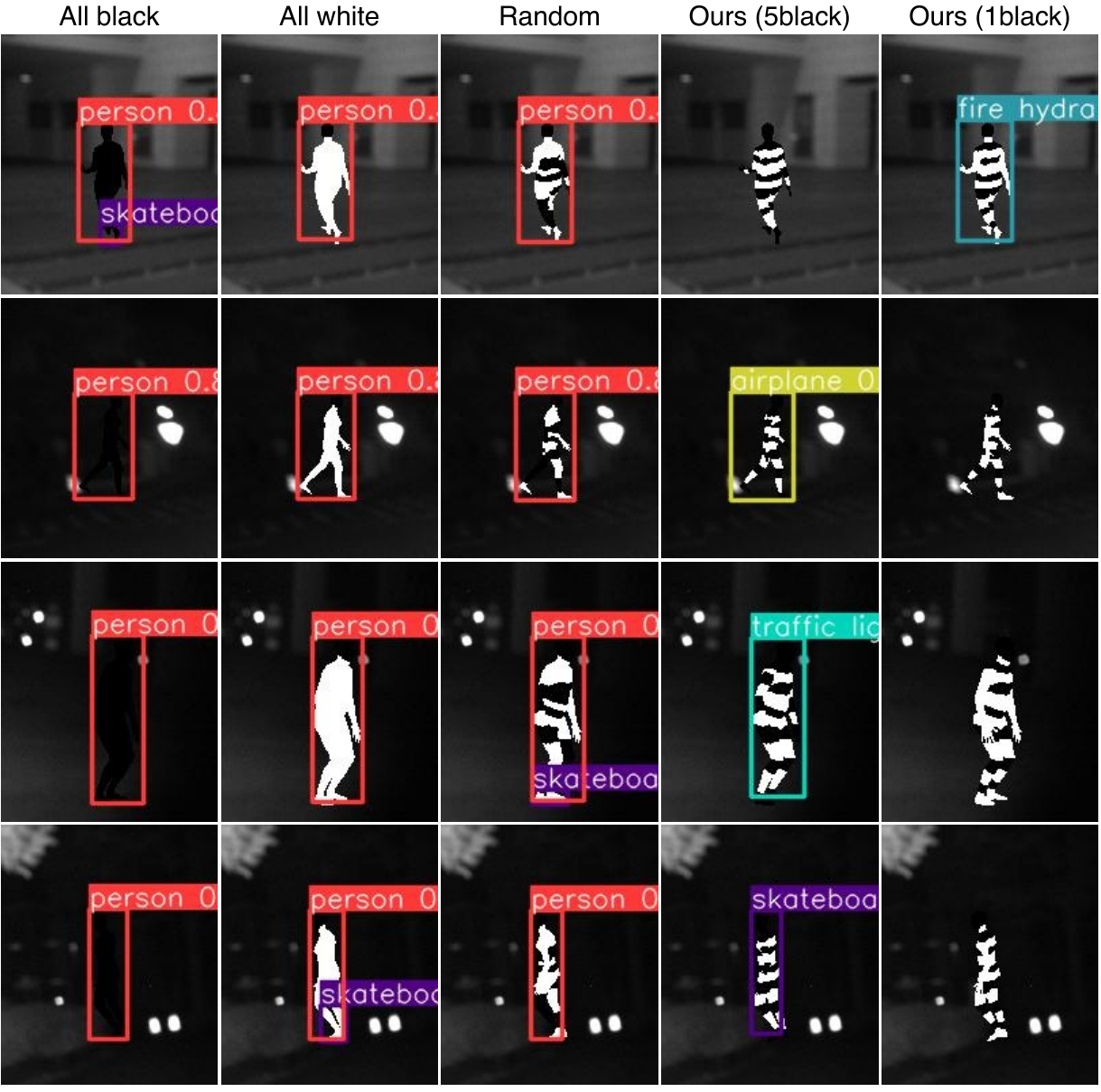}
\setlength{\abovecaptionskip}{-2mm}
\caption{Qualitative attack results of different patterns on the Official YOLO in the digital space. Images are cropped for visualization.}
\label{fig:digital}
\vspace{-4mm}
\end{figure}

\subsection{Results}

\subsubsection{Digital Attack.}

We compare five types of attack patterns based on the binary intensity manipulation, including 1) letting all body parts be black (All black), 2) letting all body parts be white except for the head (All white), 3) random assigning $0$ or $1$ to each body part, 4) using our algorithm to search the adversarial patterns with head, hands, and foots' part set to black (Ours, 5black), and 5) using our algorithm to search the adversarial patterns with head part set to black (Ours, 1black). The quantitative and qualitative results for digital attack are shown in Tab.~\ref{tab:digital} and Fig.~\ref{fig:digital}, respectively. We can see that the adversarial patterns designed by our algorithm have the best attack performance on two types of YOLO5 (Official checkpoint and finetuned checkpoint) in terms of both Average Confidence (AC) and Attack Success Rate (ASR). We can also see that after being finetuned on NIR datasets, the YOLO model is easier to attack and becomes less robust to the adversarial patterns. This verifies the severe vulnerabilities of a visual AI algorithm trained on NIR images with limited variation, as analyzed in Sec.\ref{sec:vul}.

\subsubsection{Physical Attack.}

\begin{figure}[t]
\centering
\includegraphics[width=\linewidth]{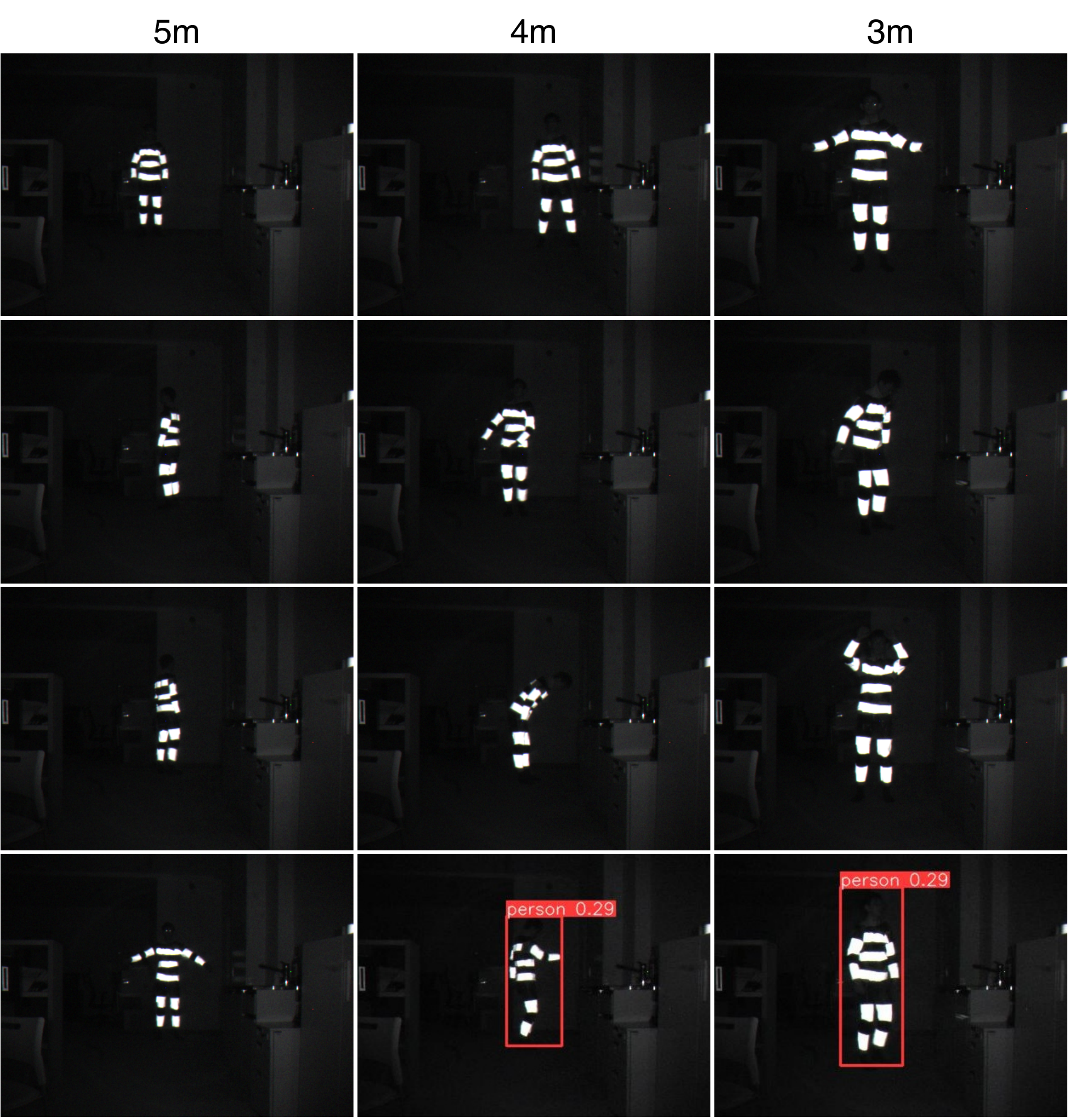}
\setlength{\abovecaptionskip}{-2mm}
\caption{Qualitative results of our designed pattern in the physical space. Images are cropped for visualization.}
\label{fig:physical}
\vspace{-2mm}
\end{figure}

\begin{figure}[t]
\centering
\includegraphics[width=\linewidth]{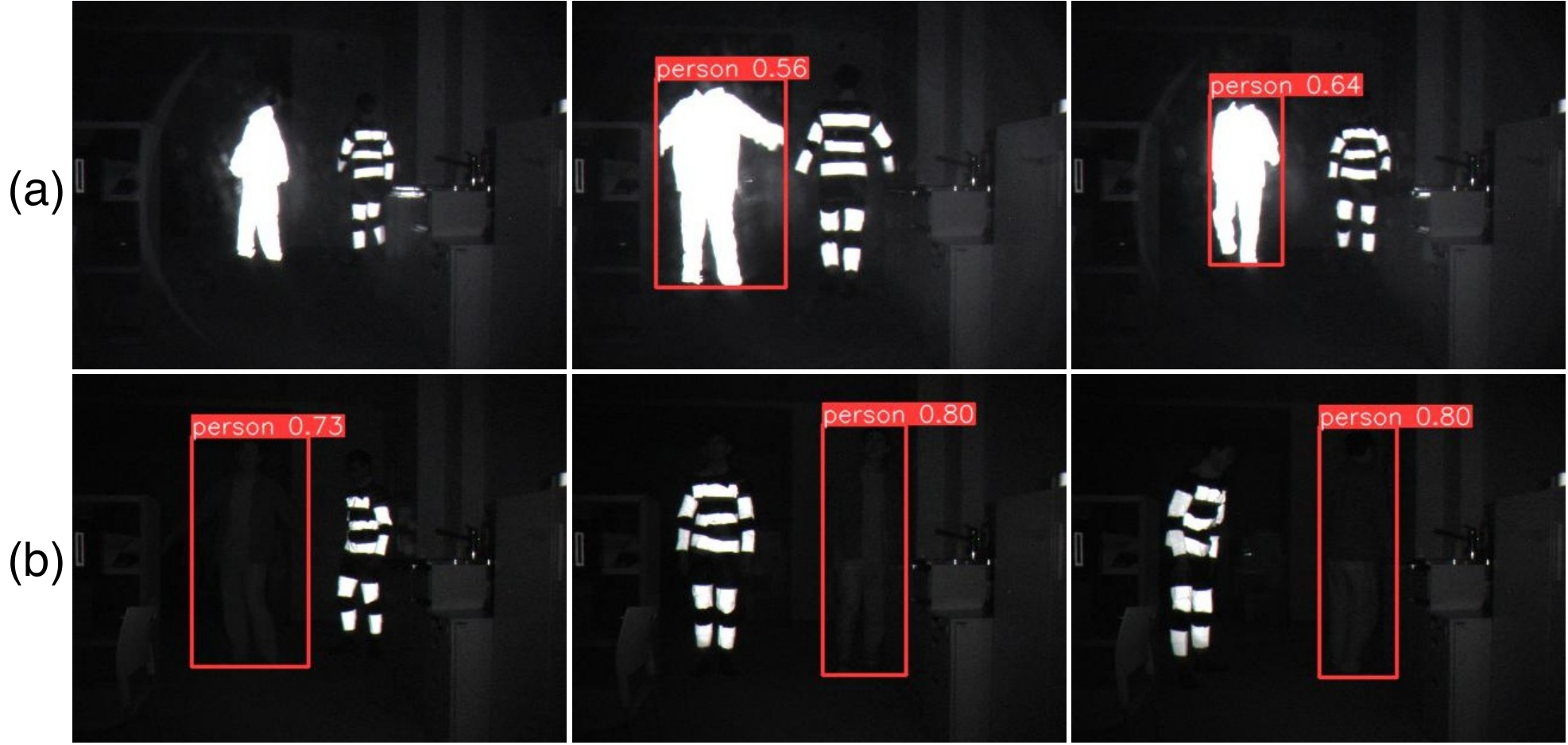}
\setlength{\abovecaptionskip}{-2mm}
\caption{Qualitative comparisons between our designed patterns, all-white patterns, and no adversarial pattern in the physical space. (a) Our designed patterns and all-white patterns, (b) Our designed patterns and no adversarial pattern. Images are cropped for visualization.}
\label{fig:physical_compare}
\vspace{-2mm}
\end{figure}

To further justify the effectiveness of our proposed attack method, we physically realize our adversarial patterns in the physical space by using the retro-reflective tapes (\ie, white) and the electrical insulating tapes (\ie, black), and use them to attack the official YOLO. During the physical attack process, we take pictures from 360 degrees of the human body and various distances to justify the effectiveness of our 3D-aware design pipeline. Specifically, we perform two groups of experiments: 1) We respectively take pictures from various angles and distances of one person with no adversarial pattern, all-white adversarial patterns, and our designed adversarial patterns. 2) We simultaneously take pictures of groups of persons with different adversarial patterns using the same camera, including one person with no adversarial pattern, one person with the all-white adversarial patterns, and one person with our designed adversarial patterns. While being captured, each person displays distinct actions instead of merely remaining stationary. The results are reported in Tab.~\ref{tab:physical}, Fig.~\ref{fig:physical}, and Fig.~\ref{fig:physical_compare}.

From Tab.~\ref{tab:physical}, we can see that our designed patterns achieve significant attack performances in the physical space, with an average Average Confidence (AC) of 0.088 and an average Attack Success Rate (ASR) of 87.93\%. Corresponding qualitative results are shown in Fig.~\ref{fig:physical}. We also analyze the impact of distances on the physical attack by collecting adversarial samples of different distances. We can see that both AC and ASR become worse as the distance of the person becomes closer to the camera. The reasons are two folds: 1) the pixel values of electrical insulating tapes no longer remain 0 when the distance becomes too close, corrupting the adversarial patterns. 2) the human head, hands and feet becomes more visible when the distance becomes very close. Considering the fact that surveillance cameras are typically installed at a considerable distance from the monitored area, we can infer that our model is relevant and meaningful for practical surveillance scenarios.

\begin{table}[!t]
\centering
\setlength{\abovecaptionskip}{2mm}
\caption{Average Confidence (AC) and Attack Success Rate (ASR) ($\%$) results of different distances in physical space. For ``No Attack'', we use Detection Rate (DR) instead of ASR. Best results are in \textbf{Bold}.}
\resizebox{\linewidth}{!}{
\begin{tabular}{lcccccc}
\toprule
\multirow{2}{*}{Distance} & \multicolumn{2}{c}{No attack} & \multicolumn{2}{c}{All white} & \multicolumn{2}{c}{Ours} \\
\cmidrule(r){2-3} \cmidrule(r){4-5} \cmidrule(r){6-7}
      & AC & DR & AC $\downarrow$ & ASR $\uparrow$ & AC $\downarrow$ & ASR $\uparrow$ \\ 
\midrule
3m     & 0.888 & 100.0 & 0.411 & 32.64 & \textbf{0.152} & \textbf{76.22} \\
4m     & 0.827 & 98.78 & 0.373 & 33.65 & \textbf{0.075} & \textbf{91.20} \\
5m     & 0.715 & 96.30 & 0.439 & 28.27 & \textbf{0.038} & \textbf{96.36} \\
\midrule
Average     & 0.810 & 98.36 & 0.408 & 31.52 & \textbf{0.088} & \textbf{87.93} \\
\bottomrule
\end{tabular}
}
\label{tab:physical}
\vspace{-2mm}
\end{table}

Fig.~\ref{fig:physical_compare} further shows the comparison between different patterns. We can see that the detector successfully detects the person without the adversarial patterns in the picture, but fail to detect the person with our designed adversarial patterns. We can also see that our designed patterns are more effective than all-white patterns in attacking the detector, which can also be proved by Tab.~\ref{tab:physical}.


\subsubsection{Evaluation of stealthiness.} 
\begin{figure}[t]
\centering
\includegraphics[width=\linewidth]{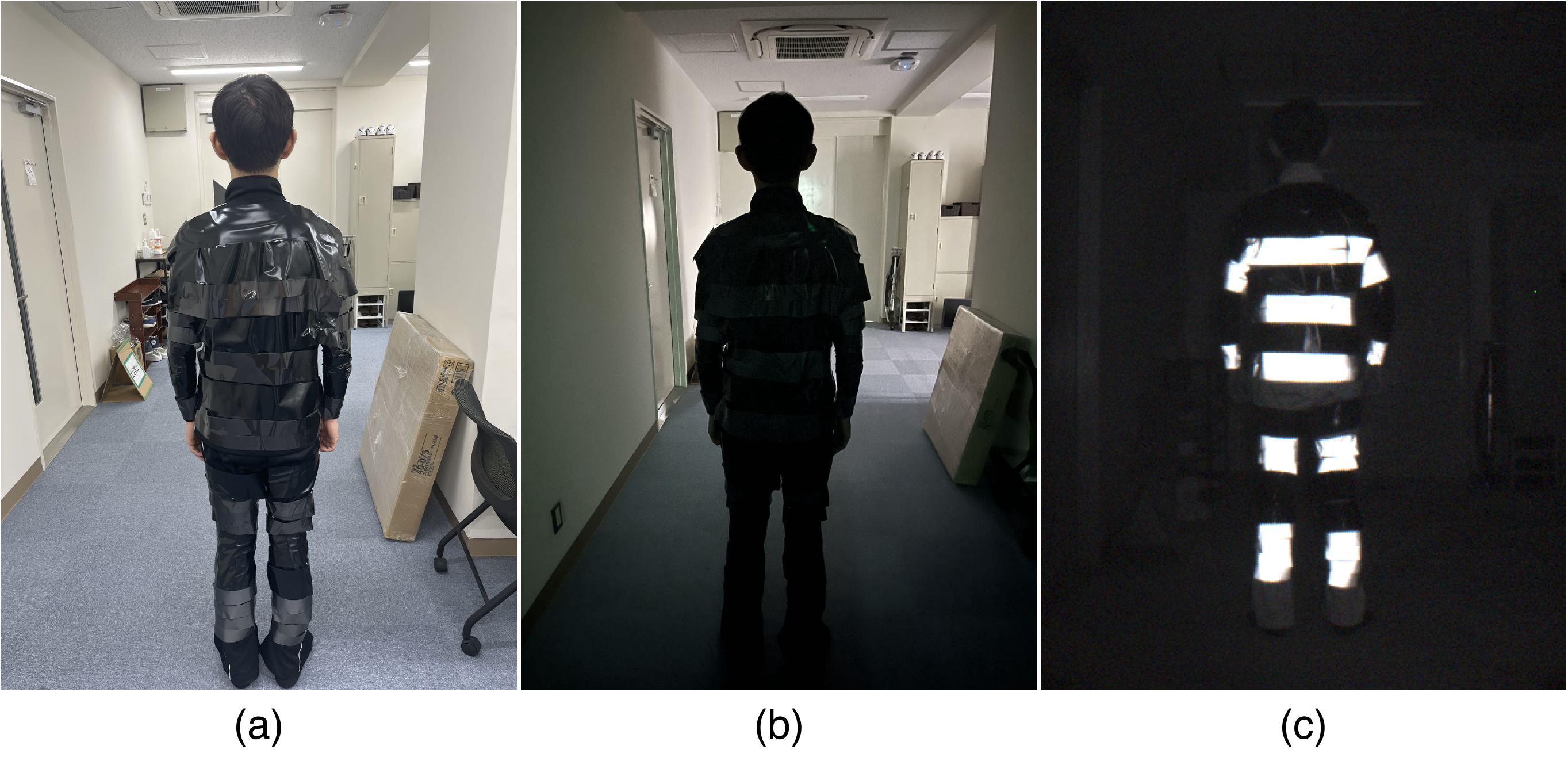}
\setlength{\abovecaptionskip}{-2mm}
\caption{Examples of RGB-NIR images with adversarial patterns. (a) RGB image under normal light, (b) RGB image under low light, (c) NIR image under co-located 850nm LED.}
\label{fig:stealthiness}
\vspace{-4mm}
\end{figure}

We choose retro-reflective tapes and electrical insulating tapes as our adversarial medium. One extra benefit is that the performance of these tapes is independent of their visible color, and even black retro-reflective tape can equally brighten NIR images, allowing the implementation of physical attackers with a natural and coherent appearance to human eyes. As shown in Fig.~\ref{fig:stealthiness}, it is clear that the adversarial mediums are in harmony with clothes and do not draw attention, especially in night environment.

\section{Discussions}

\noindent \textbf{Research ethics.}
In line with existing researches on adversarial attacks, our motivation for this study is to arouse due attention to the reliability of NIR-based AI, which has been deployed in commercial surveillance systems. We hope countermeasures to detect such attacks, through data filtering, enhancement, and more complicated strategies, could be proposed for better reliability.

\noindent \textbf{Defense methods.}
By fine-tuning the detectors on datasets containing our adversarial patterns, they are expected to become more robust against our attack. Additionally, another fundamental defense strategy involves altering the co-located setting of the surveillance system. 
Without the co-located setting, our tapes would fail to produce `white' adversarial patterns, resulting in unsuccessful attacks. However, changing the co-located setting introduces various challenges, including occlusion and increased space occupation.

\noindent \textbf{Limitations and future works.}
Although our work achieves significant attack performances in both digital and physical space, our model cannot faithfully model the head, hands, and feet texture of human bodies in the NIR domain, which leads to failure cases at very near distances. In future work, we aim to resolve this problem by collecting NIR human skin texture for more realistic rendering.

\section{Conclusion}
In this paper, we have revealed the fundamental vulnerabilities of NIR-based AI that are widely deployed in surveillance camera systems for daily use. Due to the characteristics of the spectral response of silicon sensors and the reflectance of dyed clothing, images captured in the NIR range suffer from color loss and texture loss. The special geometric configuration of the illuminant and the camera in a nearly co-located setup further allows manipulation of image intensity easily in the physical world, by using retro-reflective tapes and insulating tapes. We showed the success of attacking a YOLO-based human detector by designing an adversarial attacker in the digital space using 3D-aware techniques and later implementing it in the physical world, which arouses concerns about the reliability of surveillance systems powered by NIR AI algorithms.

\section*{Acknowledgements}
This work was partially supported by JSPS KAKENHI Grants 21H04907, 22H00529, 20H05951, JST CREST Grants JPMJCR18A6, JPMJCR20D3, and ROIS NII Open Collaborative Research 2023-23S1201. 

\bibliographystyle{ACM-Reference-Format}
\bibliography{sample-base}


\end{document}


\title{Physics-Based Adversarial Attack on Near-Infrared Human Detector for Nighttime Surveillance Camera Systems \\ SUPPLEMENTARY MATERIALS}

\maketitle

\section{Detection Threshold}

During all our experiments (both digital space and physical space), we use the threshold of $0.25$ for YOLO, following its official threshold setting.

\section{Video Experimental Results}

We further provide video results of our experiments, the corresponding videos can be found in the supplementary materials. Note that the first 9 videos are of the same person to eliminate any influence of identities on detection. The last two videos are of two distinct persons.

\begin{itemize}
    \item `1\_normal\_3m.mp4': Detection results for the person with no adversarial pattern at the distance of 3 meters.
    \item `2\_normal\_4m.mp4': Detection results for the person with no adversarial pattern at the distance of 4 meters.
    \item `3\_normal\_5m.mp4': Detection results for the person with no adversarial pattern at the distance of 5 meters.
    \item `4\_white\_m.mp4': Detection results for the person with all-white pattern at the distance of 3 meters.
    \item `5\_white\_4m.mp4': Detection results for the person with all-white pattern at the distance of 4 meters.
    \item `6\_white\_5m.mp4': Detection results for the person with all-white pattern at the distance of 5 meters.
    \item `7\_ours\_3m.mp4': Detection results for the person with our designed patterns at the distance of 3 meters.
    \item `8\_ours\_4m.mp4': Detection results for the person with our designed patterns at the distance of 4 meters.
    \item `9\_ours\_5m.mp4': Detection results for the person with our designed patterns at the distance of 5 meters.
    \item `10\_together\_normal.mp4': Detection results for two persons respectively with no adversarial pattern and our designed patterns.
    \item `11\_together\_white.mp4': Detection results for two persons respectively with all-white adversarial pattern and our designed patterns.
\end{itemize}
\bibliographystyle{ACM-Reference-Format}
\bibliography{sample-base}
